\pdfoutput=1

\documentclass[11pt]{article}

\usepackage{emnlp2021}

\usepackage{times}
\usepackage{latexsym}

\usepackage[T1]{fontenc}

\usepackage[utf8]{inputenc}

\usepackage{microtype}

\usepackage[T1]{fontenc}
\usepackage{type1cm}
\usepackage[american]{babel}
\usepackage{xurl}
\usepackage{amssymb} 
\usepackage{amsmath}
\usepackage{multirow} 
\usepackage{xspace}
\DeclareMathAlphabet{\mathcalbf}{OMS}{pzc}{b}{n}
\usepackage{microtype}
\usepackage{bm}
\usepackage{siunitx}

\usepackage[pdftex]{graphicx}
\usepackage{tabularx}
\usepackage{booktabs}
\usepackage{subcaption}
\DeclareGraphicsExtensions{.pdf,.ai,.jpg,.png}
\setkeys{Gin}{pagebox=artbox}

\graphicspath{{../argmining21-sufficiency-generation-figures/}}

\newcommand{\bsfigure}[3][]{%
	\begin{figure}[t]
		\centering
		\includegraphics[#1]{#2}
		\caption{#3}\label{#2}%
	\end{figure}
}

\RequirePackage{type1cm}
\RequirePackage{color}
\RequirePackage{soul}
\setstcolor{blue}
\definecolor{lightgray}{rgb}{0.95,0.95,0.95}
\definecolor{lightgreen}{rgb}{0.56,0.93,0.56}
\definecolor{lightblue}{rgb}{0.3,0.3,0.9}
\definecolor{tgray}{rgb}{0.5,0.5,0.5}

\newsavebox\bscombox
\newcommand{\bscom}[3][]{%
	\sbox{\bscombox}{\fontsize{8}{9}\selectfont#1#2#3}
	\noindent
	\st{#2}{\selectfont
		\color{blue}#3\ifx\\#1\\\else{\fontsize{8}{9}\selectfont\color{violet}[#1]}\fi
	}
}

\begin{document}

%
%

\title{Assessing the Sufficiency of Arguments through Conclusion Generation}


\author{
  	Timon Gurcke \quad Milad Alshomary \quad Henning Wachsmuth \\
  	Department of Computer Science\hspace*{1em} \\
	Paderborn University, Paderborn, Germany \\
	{\tt<first name>.<last name>@uni-paderborn.de} \\
}

\date{}
\maketitle

\begin{abstract}
The premises of an argument give evidence or other reasons to support a conclusion. However, the amount of support required depends on the generality of a conclusion, the nature of the individual premises, and similar. An argument whose premises make its conclusion rationally worthy to be drawn is called \textit{sufficient} in argument quality research. Previous work tackled sufficiency assessment as a standard text classification problem, not modeling the inherent relation of premises and conclusion. In this paper, we hypothesize that the conclusion of a sufficient argument can be {\em generated} from its premises. To study this hypothesis, we explore the potential of assessing sufficiency based on the output of large-scale pre-trained language models.
Our best model variant achieves an F$_1$-score of .885, outperforming the previous state-of-the-art and being on par with human experts. While manual evaluation reveals the quality of the generated conclusions, their impact remains low ultimately.
\end{abstract}

\section{Introduction}
\label{sec:introduction}

The quality assessment of natural language argumentation is nowadays studied extensively for various genres and text granularities, from entire news editorials~\cite{elbaff:2020} to arguments in online forums~\cite{lauscher:2020} to single claims in social media discussions~\cite{skitalinskaya:2021}. The reason lies in its importance for  driving downstream applications such as writing support \cite{stab2017essaycorpus}, argument search~\cite{wachsmuth2017argsme}, and debating technologies~\cite{slonim:2021}. \newcite{wachsmuth2017qualitydimensions} organized quality dimensions of arguments into three complementary aspects: logic, rhetoric, and dialectic. Logical quality refers to the actual argument structure, that is, how strong an argument is in terms of the support of a claim (the argument's {\em conclusion}) by evidence and other reasons (the {\em premises}). 

\bsfigure{example-arguments}{Two example arguments from a persuasive student essay, one classified as {\em sufficient}, the other as {\em insufficient} in the corpus of \newcite{stab2017recognizingsupportedarguments}.}

A key dimension of logical quality is {\em sufficiency}, capturing whether an argument's premises together make it rationally worthy of drawing its conclusion \cite{johnson2006logical}. Consider, for example, the two arguments on health education in Figure~\ref{example-arguments}, taken from the argument-annotated essay corpus of \newcite{stab-gurevych-2017aae2}. While the upper one was deemed sufficient by human experts \cite{stab2017recognizingsupportedarguments}, the lower one was not, likely because the second premise tries to reason from a single example. A reliable computational assessment of argument sufficiency would allow systems to determine those arguments that are well-reasoned.

As detailed in Section~\ref{sec:related_work}, previous approaches to argument sufficiency assessment model the task as a standard text classification problem and tackle it with convolutional neural networks \cite{stab2017recognizingsupportedarguments} or traditional feature engineering \cite{wachsmuth:2020}. In the focused domain of persuasive student essays, \newcite{stab2017recognizingsupportedarguments} obtained a macro F$_1$-score of .827, not far away from human performance in their setting (.887). However, to further improve the state of the art, we expect the integration of knowledge beyond what is directly available in the text at hand is needed. In particular, we observe that existing work neither explicitly considers an argument's premises and conclusions, nor a property of their relationship. We hypothesize that only a sufficient argument makes it possible to infer the conclusion from the premises. Consequently, comparing the stated conclusion of an argument with one that is (automatically) {\em generated} from the premises could help the model to distinguish sufficient arguments from insufficient ones. This hypothesis raises the question of whether the knowledge encoded in large-scale pre-trained language models can be leveraged, a direction nearly unexplored so far in argument quality assessment.

In this paper, we study whether generating a conclusion from an argument's premises benefits the computational assessment of the argument's sufficiency. In particular, we first enrich the argument with structural annotations, highlighting which parts are the premises and which part is the conclusion. We propose in Section~\ref{sec:approach} to then mask the conclusion in order to learn to re-generate it using fine-tuned BART \cite{lewis2019bart}. Combining the generated conclusion with the original argument and its annotations, our approach learns to distinguish sufficient from insufficient arguments using a modified RoBERTa model \cite{liu2019roberta}.

Starting from ground-truth argument structure, we subsequently evaluate conclusion generation and sufficiency assessment on the merged annotations of the corpora of \newcite{stab-gurevych-2017aae2} and \newcite{stab2017recognizingsupportedarguments}, as described in Section~\ref{sec:data}. Our generation experiments indicate that fine-tuning BART leads to better conclusions, which are on par with human-written conclusions in terms of sufficiency, likeliness, and novelty (Section~\ref{sec:experiments}). To quantify the impact on sufficiency assessment, we explore various combinations of premises, original conclusion, and generated conclusion in systematic ablation tests, and we compare them to the state of the art and a human upper bound (Section~\ref{sec:assessment}). Our sufficiency experiments reveal that, even on the plain input text of an argument, RoBERTa already improves significantly over the state of the art. The addition of structural annotations and the generated conclusion lead to further improvements, although the benefit of generation ultimately remains limited, possibly due to the generally limited importance of knowing the conclusion on the given data.
Finally, we discuss the results of our approaches in Section \ref{sec:discussion} in light of their implications for the field.

The main contributions of this paper are:%
\footnote{The experiment code can be found under: \url{https://github.com/webis-de/ArgMining-21}}
\begin{itemize}
	\setlength{\itemsep}{0pt}
	\item
	A language model that can generate human-like argument conclusions.
	\item
	The new state-of-the-art approach to argument sufficiency assessment.
	\item
	Insights into the importance of mined and generated structure within argument assessment.
\end{itemize}
\section{Related Work}
\label{sec:related_work}

Computational argumentation research has assessed various dimensions of argument quality. \newcite{wachsmuth2017qualitydimensions} provide a theory-based taxonomy of 15 logical, rhetorical, and dialectical quality dimensions and of the work in natural language processing done in these directions. We focus on the (local) {\em sufficiency} dimension, which is key to logical cogency, representing that an argument's conclusion can rationally be drawn from its premises, given that these are acceptable and relevant \cite{johnson2006logical}.

Few approaches tackled sufficiency computationally so far. Aside from \newcite{wachsmuth:2020} who assess it as one of the 15 dimensions above using traditional text-focused feature engineering, we are only aware of the work of \newcite{stab2017recognizingsupportedarguments} who extend the argument-annotated essay corpus of \newcite{stab-gurevych-2017aae2} with binary sufficiency annotations. On this basis, the authors compare a support vector machine using lexical, syntactic, and length features to a convolutional neural network (CNN) with word vectors, the latter achieving the best result with a macro F$_1$-score of .827. In our experiments, we use their dataset and replicate their experiment settings, in order to compare to the CNN.

Unlike \newcite{stab-gurevych-2017aae2}, we rely on a transformer-based architecture, namely we adapt RoBERTa \citep{liu2019roberta} to assess sufficiency.  Approaches to argument quality assessment using such architectures are still limited, mostly focusing on a holistic view of quality \citep{gretz2019ibmrankbert, toledo2019automatic}, although a few approaches used transformers for some of the dimensions of \newcite{wachsmuth2017qualitydimensions}, such as \newcite{lauscher:2020}, or somewhat related dimensions in light of quality improvement \cite{skitalinskaya:2021}.

In contrast to the standard use of transformers for text classification, we leverage the structure of arguments for their assessment. \citet{wachsmuth2016using} provided evidence that mining the argumentative structure of persuasive essays helps to better assess four essay-level quality dimensions of argumentation. Similarly, we use annotations of the premises and conclusions of arguments for the sufficiency assessment, but we target the arguments. Moreover, we explore to benefit of conclusion {\em generation} for the assessment.

The idea of reconstructing an argument's conclusion from its premises was introduced by \newcite{alshomary2020target}, but their approach focused on the inference of a conclusion's target. The actual generation of entire conclusions has so far only been studied by \citet{syed2021generating}. The authors presented the first corpus for this task along with experiments where they adapted BART \citep{lewis2019bart} from summarization to conclusion generation. While they  trained BART to directly generate a conclusion based on premises, we generate conclusions that fit the context of an entire argument. To this end, we leverage and finetune BART's inherent denoising capabilities obtained during pre-training to replace a mask token in an argument.

\section{Data}
\label{sec:data}

To study our hypothesis that a sufficient argument's conclusion can be generated from its premises, we need data that is annotated for both argument structure and sufficiency. In this section, we describe how we employ existing corpora for this purpose.

\subsection{Data for Conclusion Generation}
\label{sec:data:generation}

The argument-annotated essay (AAE-v2) corpus \citep{stab-gurevych-2017aae2} contains structural annotations for 402 complete persuasive student essays. For conclusion generation (as well as for structure-based sufficiency assessment), we only need annotations of single arguments.

Our instance creation procedure resembles the one of \newcite{alshomary2020target}, but we work on argument level rather than conclusion level, since we approach conclusion generation as a language model denoising task \citep{lewis2019bart}. Concretely, instead of using premises-conclusion training pairs where the conclusion shall be generated given the premises, we rely on argument-argument pairs. The first argument here is a modified version of the second argument where the conclusion is masked. This way, we avoid conflicts with the argument-level sufficiency annotations of \citet{stab2017recognizingsupportedarguments} (see below). In total, we obtain 1506 argument-argument pairs relating to 1506 unique conclusions matched with 1029 unique arguments. On average, each argument has a length of 4.5 sentences and contains 94.6 tokens. 

For training, we rely on 5-fold cross-validation, ensuring that the argument-argument pairs from one essay are never split between training, validation, and test data. This prevents possible data leakage that could artificially improve the final evaluation scores. For each folding, we use 70\% training, 10\% validation, and 20\% test data.

\subsection{Data for Sufficiency Assessment}
\label{sec:data:sufficiency}

\citet{stab2017recognizingsupportedarguments} further classified each argument in the 402 essays of the AAE-v2 corpus as being sufficient or not. Following \citet{johnson2006logical}, the authors defined that an ``argument complies with the sufficiency criterion if its premises provide enough evidence for accepting or rejecting the claim'' (we speak of ``conclusion'' here instead of ``claim''). All 1029 arguments were labeled, of which 681 (66.2\%) were considered sufficient and 348 (33.8\%) insufficient.

We use the provided corpus both in its original form and in a modified version where we replace the conclusion of an argument with two separator tokens, ``</s></s>''. This allows us to study a wide range of different approaches for sufficiency assessment by placing text in-between the two tokens as a replacement for the original conclusion.

For training, we replicate the original 20-times 5-fold cross-validation setup of \citet{stab2017recognizingsupportedarguments}, with 70\% training, 10\% validation, and 20\% test data, in order to ensure comparability.

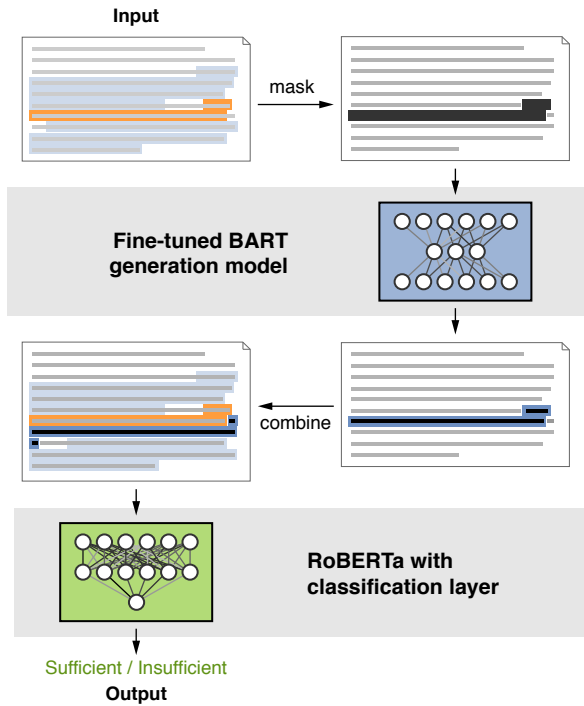
\bsfigure{approach}{Illustration of our sufficiency assessment approach through generation: (1) BART is used to generate the masked conclusion in an argument. (2) The generated conclusion is combined with the ground truth annotation of the argument. (3) RoBERTa classifies the enriched argument as sufficient/insufficient. Several ablations of the annotations are tested in our experiments.}

\section{Approach}
\label{sec:approach}

This section describes our two-step approach to assess the sufficiency of a given argument through conclusion generation. First, we generate a conclusion from the argument's premises using a pre-trained language model finetuned on the task of replacing the masked conclusion of an argument. Second, the generated conclusion is used to assess the argument's sufficiency by experimenting with eight modified versions of the original input argument (Section~\ref{sec:assessment}). An overview of the approach is shown in Figure~\ref{approach}. In the following, we detail how we train the models for the two steps.

\subsection{Conclusion Generation using Denoising}

Given an argument with a masked conclusion, the first task is to re-generate the conclusion. To tackle this task, we use BART-large \citep{lewis2019bart} and treat generation as a denoising task. We explore two model variants:

\paragraph{BART-unsupervised}

In this variant, we do not finetune BART on any data, but we use its vanilla denoising capabilities obtained in its pre-training procedure. Note that the masked conclusions usually do not represent entire sentences, thus leaving textual markers which trigger BART to generate a logical conclusion, for example, ``Thus, <mask>'' or ``This makes it clear that <mask>.'' We consider this model as a baseline.

\paragraph{BART-supervised}

In this variant, we finetune BART on the data from Section \ref{sec:data}, in order to tailor its denoising capabilities towards conclusion generation. In particular, we thereby adjust the language model towards the given domain and teach the model to replace the mask token with a conclusion (instead of just generating text that fits the context). We finetune BART using cross-entropy loss, as commonly done in text generation.

The proper training settings of the two models are found via a hyperparameter search. For the evaluation described below, we ran 10 trials for each fold, testing batch sizes between 4 and 8 and learning rates between $5 \cdot 10^{-6}$ and $5 \cdot 10^{-5}$. We fixed the number of epochs to 3 per fold, as we did not observe any improvements afterwards, and we used a cosine learning rate scheduler with 50 warm-up steps to stabilize the training. At inference time, we employed a beam size of 5 to obtain the final generated conclusions. We considered the epoch out of three, which performs best on the validation data in terms of BERTScore \citep{zhang2019bertscore}.

\subsection{Sufficiency Assessment using Structure}

Given a modified argument, the second task is to predict whether the premises in the argument are rationally worth drawing the conclusion. We use RoBERTa \citep{liu2019roberta} for this task by adding a linear layer on top of the pooled output of the original model \citep{devlin2018bert}.

Successfully optimizing RoBERTa using mean squared error (MSE) or cross-entropy loss functions would be difficult, as both of them do not align well with the target metric of sufficiency assessment (macro F$_1$-score). We therefore follow ideas of \citet{puthiya2014optimizing} and \citet{eban2017scalable} who propose to optimize machine learning models on the F$_1$-score directly.
Accordingly, we allow the model to output probabilities instead of interpreting a single binary value.
Analogous to \citet{stab2017recognizingsupportedarguments}, we allow our model to adjust hyperparameters between folds.
In our experiments, we followed the same hyperparameter optimization procedure as before but for different parameters and ranges. In total, we ran 10 trials for each fold, and we adjusted the batch size to be between 16 and 32 and the learning rate between $10^{-6}$ and $5 \cdot 10^{-5}$. We selected the epoch for each trial out of three, which performed best on the validation data in terms of macro F$_1$-score.

\section{Evaluation of Conclusion Generation}
\label{sec:experiments}

To study our hypothesis, we need to ensure that the generated conclusions are meaningful and fit in the context of a given argument, so they can be helpful in sufficiency assessment. In this section, we therefore evaluate the quality of the generated conclusions, both automatically and manually.

\subsection{Automatic Evaluation}

\begin{table}[tb]
\centering
\small
\setlength{\tabcolsep}{3pt}
\begin{tabular}{l@{\hspace*{-1.5em}}rrrr}
\toprule
\bfseries{Model}  & \bfseries{BERTScore} & \bfseries{ROU.-1} & \bfseries{ROU.-2} & \bfseries{ROU.-L}  \\
\midrule
BART-unsupervised   & 0.14          & 19.69          & \phantom{0}4.05 & 16.40          \\
BART-supervised     & \bf 0.25          & \bf20.97          & \bf \phantom{0}4.79 & \bf 17.49          \\
\bottomrule
\end{tabular}
\caption{Automatic evaluation results of concluion generation: Rescaled F1-BERTScore and ROUGE-1/-2/-L scores of the two considered models on the full corpus.}
\label{table-automatic-evaluation-internal}
\end{table}



As indicated in Section~\ref{sec:approach}, we compare two approaches:
(1) {\em BART-unsupervised}, which replaces the mask token in an argument (denoising) with fitting text, as it is part of BART's training procedure \cite{lewis2019bart}; and 
(2) {\em BART-supervised}, which finetunes BART on the argument-argument pairs from Section \ref{sec:data}.
For both approaches, we obtained the complete set of generated conclusions using the cross-validation setup described in Section~\ref{sec:approach}. Matching these with the corresponding ground-truth conclusion, we then computed their quality in terms of BERTScore as well as ROUGE-1, ROUGE-2, and ROUGE-L.

\paragraph{Results}

Table \ref{table-automatic-evaluation-internal} lists the results of the approaches. BART-unsupervised is a strong baseline in terms of lexical accuracy: Values such as 19.69 (ROUGE-1) and 16.40 (ROUGE-L) are comparable to those that \newcite{syed2021generating} achieved in similar domains with sophisticated approaches. However, finetuning on the argument-argument pairs does not only significantly increase the semantic similarity between generated and ground truth conclusions from 0.14 to 0.25 in terms of BERTScore, but it also leads to a slight increase in lexical accuracy.

\subsection{Manual Evaluation}

\begin{table*}[t!]
\small
\setlength{\tabcolsep}{4pt}
\renewcommand{\arraystretch}{1.0}
\centering
\begin{tabular*}{\linewidth}{lllcrrrrrcrrcrr}
\toprule
	&							&					&& \multicolumn{2}{c}{\bf (a) Agreement}	&& \multicolumn{5}{c}{\bf (b) Majority Scores}	&& \multicolumn{2}{c}{\bf (c) Mean Results} \\
						   							\cmidrule{5-6}		      				\cmidrule{8-12}	  							\cmidrule{14-15}	 
\bf \# & \bf Question 					& \bf Approach 			&& $\alpha$ 	& \bf Majority			&& \bf 1 	& \bf 2 	& \bf 3 	& \bf 4 	& \bf 5 	&& \bf Score  $\uparrow$ 	& \bf Rank $\downarrow$  \\
\midrule
Q1	& Are the premises sufficient 		& Ground Truth			&& .23 		& \bf 63\%				&& 7         & 24		& 39    	& 30	    	& 0		&& \bf 2.92		& 1.43		\\
	& to draw the conclusion?			& BART-unsupervised	&& \bf.43		& \bf 63\%				&& 14	& 19	    	& 34		& 32    	& 1		&& 2.87			&1.40		\\
	&							& BART-supervised		&& .35		& 60\%	    			&& 14 	& 21	    	& 30		& 34 		& 1		&& 2.87 			&\bf 1.37		\\					
\addlinespace		
Q2	& How likely is it that the			& Ground Truth			&& .19		& 57\%	   		 	&& 4 	& 18		& 54		& 24		& 0		&& \bf 2.98 		&1.42		\\
	& conclusion will be inferred 		& BART-unsupervised	&& \bf .32		& \bf 64\%				&& 16	& 15		& 46		& 23		& 0		&& 2.76 			&1.54		\\
	& from the context?				& BART-supervised	    	&& .28		& \bf 64\%				&& 12	& 15		& 38	 	& 35		& 0		&& 2.96 			&\bf 1.39		\\ 				
\addlinespace	
Q3	& How can the conclusion be 		& Ground Truth			&& .19		& 72\%				&& 0		& 4 		& 18	    	& 73 		& 5		&& \bf 3.79 		&\bf 1.39		\\
	& composed from the context?		& BART-unsupervised	&& \bf .50		& 80\% 				&& 14	& 11		& 15		& 47	    	& 13		&& 3.34 			&1.54		\\
	&							& BART-supervised		&& .32		& \bf 82\%				&& 8	    	& 8 	    	& 22		& 53	    	& 9		&& 3.47  			&1.54		\\					
\bottomrule
\end{tabular*}
\caption{Manual evaluation results of conclusion generation on the 100 arguments of each of the three approaches: (a)~Agreement of all five annotators in terms of Krippendorff's $\alpha$ and majority. (b)~Distribution of majority scores. For Q1/Q2, higher scores mean more sufficient/likely. For Q3, they mean less ``copying'' (see text for details). (c)~Mean score of each approach and rank obtained by comparing the majority score for each argument in isolation.}
\label{table-agreement-and-correlations}
\end{table*}

The metrics used to automatically evaluate conclusion generation are not ideal, since they expect a single correct result. 
The task of conclusion generation, in contrast, allows for multiple, possibly very different correct conclusions, for example, a different conclusion target may be derived from a single set of premises \cite{alshomary2020target}. 
We thus conducted an additional manual annotation study to evaluate the quality of the conclusions generated by the two approaches in comparison to the human ground truth.

We randomly chose 100 arguments from the given corpus, 50 labeled as sufficient and 50 labeled as insufficient. For each arguments, we additionally created two variants, replacing the original conclusion with the generated conclusion of either approach. In each case, we then presented the three arguments with their premises and conclusions highlighted to five annotators of different academic backgrounds (economics, computer science, health/medicine), none being an author of this paper. We asked each annotator three questions, Q1--Q3, on each argument, resulting 300 annotations for each model and 900 annotations in total. For consistency reasons, we used a 5-point Likert scale for each question:

\begin{itemize}
\setlength{\itemsep}{0pt}
\item
Q1: Are the premises sufficient to draw the conclusion?
This question referred to the {\em sufficiency} of arguments, from ``not sufficient'' (score~1) to ``sufficient'' (score~5).
We asked this question to see how the sufficiency of generated and human-written conclusions differs, thus directly evaluating our hypothesis.
\item
Q2: How likely is it that the conclusion will be inferred from the context? This question referred to the {\em likelihood} of a conclusion, from ``very unlikely'' (score~1) to `` very likely'' (score~5). We asked this question as an internal quality assurance, ruling out the possibility that our models generate conclusions unrelated to the given context of the argument. 
\item
Q3: How can the conclusion be composed from the context? This question, finally, referred to the {\em novelty} of the generated conclusions in light of their context. The score range here is more complex; inspired by \newcite{syed2021generating}, who also ask annotators about the novelty of generated conclusions, it reflects the cognitive load required to infer the conclusions from the context of the argument:  ``verbatim copying''~(1), ``synonymous copying''~(2), ``copying + fusion''~(3), ``inference''~(4), and ``can not be composed"~(5).
\end{itemize}

\paragraph{Results}

For each question, Table \ref{table-agreement-and-correlations} shows the inter-annotator agreement, the distribution of majority scores, and the resulting mean score of the three compared approaches (including the ground truth), and the mean rank. We obtained the rank by treating each question as a ranking task where, for each argument, the approaches are ranked  from~1 to~3 by decreasing highest majority score.

We find that the general {\em agreement} for the questions in terms of Krippendorff's $\alpha$ is low, with values between .19 and .50, but comparable to other tasks in the realm of argumentation \citep{wachsmuth2017qualitydimensions}. On all three questions, the annotators agreed mostly for BART-unsupervised, followed by the BART-supervised, while having the least agreement for the ground truth. This may indicate a more apparent connection of the generated conclusions to the premises. In 57\% to 64\% of the cases, we observe majority agreement of the annotators for the first two questions, whereas this value goes up to 72\%--78\% for the last question.

The ranking for Q1 shows that conclusions generated by our BART-supervised model overall ranked best in {\em sufficiency} (mean rank 1.37), even though the mean score is slightly better for the ground truth. While the differences are small, this behavior is expected as half of the provided arguments were initially labeled as insufficient. 

Considering the {\em likelihood} of the premises~(Q2), we see that conclusions of BART-supervised are on par with the ground truth conclusions (score rank 2.96 vs.\ 2.98, rank 1.39 vs.\ 1.42) and better than those generated by the baseline BART-unsupervised (1.54). This suggests that the conclusions generated by our model both fit the context of the argument and are at least as likely to be drawn as the ones written by humans. This property is essential for our approach to sufficiency assessment to rule out the possibility of failure due to a low quality of the generated conclusions in general. 

Finally, we consider the cognitive load that is required to compose a conclusion given its context, as reflected by {\em novelty}~(Q3). This information is vital to rule out that the generated conclusions are copied from the context of an argument instead of being inferred. We find the ground-truth conclusions to require the most cognitive load in this regard, having a clearly better mean rank (1.39) than the others (both 1.54). Thus, they potentially provide the most novelty to the context. The mean scores indicate an increase of novelty from BART-unsupervised (3.34) to BART-supervised (3.47) though.

\begin{table*}[t!]%
\centering%
\small
\renewcommand{\arraystretch}{1.0}
\setlength{\tabcolsep}{2.5pt}%
\begin{tabular}{p{0.25cm}p{1.5cm}p{2.75cm}p{10.75cm}}

\toprule
\bf \#	& \textbf{Label}		& \textbf{Part} 			& \textbf{Text}  \\
\midrule
(a)	& Sufficient		& Argument 			& Second, \texttt{<MASK>}. Averagely, public transports use much less gasoline to carry people than private cars. It means that by using public transports, the less gas exhaust is pumped to the air and people will no longer have to bear the stuffy situation on the roads, which is always full of fumes.	\\
\addlinespace
	&				& Ground truth	 		& public transportation helps to solve the air pollution problems	  \\
\addlinespace
	&				& BART-unsupervised	& public transport is more efficient than private cars 		  \\
\addlinespace
	&				& BART-supervised		& using public transports will help to reduce the amount of pollution in the air \\
\midrule
(b)	& Insufficient		& Argument 			& Last, \texttt{<MASK>}. Playing musical instrument is a good way, I can play classical guitar. When I meet difficulties in studies, I will take my guitar and play the song Green Sleeves. It makes me feel better and gives me the confidence.	\\
\addlinespace
	&				& Ground truth	 		& we should develop at least one personal hobby, not to show off, but express our emotion when we feel depressed or pressured	  \\
\addlinespace
	&				& BART-unsupervised	& but not least, I love music 		  \\
\addlinespace
	&				& BART-supervised		& playing musical instrument is very important to me \\
%
\midrule
(c)	& Insufficient		& Argument 			& In addition to this, \texttt{<MASK>}. For instance, further enforcement banned smoking in capital in Sri Lanka has reduced this consumption related diseases and deaths, as per the ministry of health. As this shows, smoking restrictions has successfully daunted public from this bad puffing that put less strain on country's healthcare systems.	\\
\addlinespace
	&				& Ground truth	 		& introducing smoking ban in public places would greatly discourage people from engaging tobacco puffing	  \\
\addlinespace
	&				& BART-unsupervised	& Sri Lankan government has taken several measures to curb smoking		  \\
\addlinespace
	&				& BART-supervised		& smoking restrictions in Sri Lanka has brought a lot of benefits to the country \\
\bottomrule
%
%
\end{tabular}%
\caption{Conclusions generated by the BART models for four arguments (with masked conclusion) compared to the {\em ground truth} conclusion: (a)~BART-supervised almost reconstructs the ground truth. (b)~Here, the two models increase sufficiency. (c)~Sometimes, the generated conclusions remain rather vague and pick the wrong target.}
\label{table-examples}%
\end{table*}

The scores and ranks can be interpreted more easily when looking at the {\em majority scores} for the two BART models and the ground truth. As expected, we find that the amount of conclusions considered to be sufficient is approximately the same to those considered insufficient. The annotators did not agree on the highest  sufficiency rank, which may be due to subjectivity in the perception of ``full'' sufficiency. We observe an analog behavior for  score~5 for the likelihood of conclusions (Q2). Here, this may imply that, rarely, only a single conclusion would fit the premises of an argument. Regarding the novelty of the generated conclusions, Q3 reveals that the ground truth annotations are mostly inferences (score~4) and only rarely copied from the context (scores~1 and~2). While BART-unsupervised only somewhat follows this distribution, our BART-supervised model shows a similar behavior to the ground truth, though in less clear form. This is another indication that BART-supervised has learned to generate conclusions from premises.

\paragraph{Error Analysis \& Examples}
To better understand the differences between the two BART models and the ground truth, we analyzed the 100 examples from our annotation study manually. Table~\ref{table-examples} compares the conclusions for three arguments. 

For the sufficient argument in Table~\ref{table-examples}(a), BART-supervised nearly perfectly reconstructs the ground truth, whereas BART-unsupervised generates a reasonable but less specific alternative. In Table~\ref{table-examples}(b), the approaches appear to make the argument more sufficient, particularly BART-supervised. Both examples speak for the truth of our hypothesis that the conclusion of sufficient arguments can be generated from the premises. This tendency is further backed by our analysis in which we found that for BART-supervised 20\% (10/50) of the sufficient arguments have perfect matching conclusions (Table~\ref{table-examples}(a)) 60\% (30/50) either are less abstract or have a different conclusion target but are of equal quality, and 20\% (10/50) have a conclusion that is of lower quality, compared to the ground truth. In contrast, only 10\% (5/50) of the insufficient arguments are perfect matches, 60\% (30/50) either are less abstract or have a different conclusion target of equal quality, and 30\% (15/50) have lower quality generated conclusions (Table~\ref{table-examples}(b)).

Further analysis reveals that for 58\% (29/50) of the insufficient arguments, BART-supervised generated a sufficient conclusion with a different target (16/29) or a different level of abstraction, either more specific (11/29) or more abstract (2/29) than the ground truth. In Table~\ref{table-examples}(c), the BART models seem tricked by the anecdotal evidence given in the premises, mistakenly picking Sri Lanka as the conclusion's target.

Regarding the difference of our models, we found that BART-supervised more often generated conclusions with targets that are equally likely to the target of the human ground truth (68/100 vs. 42/100).

\begin{table*}[t!]
\small
\setlength{\tabcolsep}{7pt}
\renewcommand{\arraystretch}{1}
\centering
\begin{tabular*}{0.95\linewidth}{ll r@{\;}l r@{\;}l r@{\;}l r@{\;}l}
\toprule
\bf Assessment &	\bf Approach 					& \multicolumn{2}{c}{\bf Accuracy} & \multicolumn{2}{c}{\bf Macro Pre.} & \multicolumn{2}{c}{\bf Macro Rec.}	& \multicolumn{2}{c}{\bf Macro F1}  \\
\midrule
Direct		& Human upper bound				& .911 	& $\pm$ .022      	& .873 	& $\pm$ .042       	& .903  	& $\pm$ .020      & .883 	& $\pm$ .029           				 \\
\addlinespace
			& CNN	 \cite{stab-gurevych-2017aae2}	& .846 	& $\pm$ .022      	& .830 	& $\pm$ .021       	& .832  	& $\pm$ .028      & .831 	& $\pm$ .023           				 \\
			& \bf RoBERTa						& \bf .889 	& $\pm$ .026      	& \bf .882 	& $\pm$ .057       	& \bf .880  & $\pm$ .078      & \bf .876 & $\pm$ .031$^{\dagger}$           	 \\
\midrule
Indirect		& RoBERTa-premises-only			& .887 	& $\pm$ .031      	& .876 	& $\pm$ .051       	& .881 	& $\pm$ .071 	& .875 	& $\pm$ .037           				  \\
			& RoBERTa-conlusion-only			& .641 	& $\pm$ .036      	& .582 	& $\pm$ .048       	& .567 	& $\pm$ .144	& .553 	& $\pm$ .063           				  \\
			& RoBERTa-generated-only			& .632 	& $\pm$ .025      	& .560 	& $\pm$ .038       	& .544 	& $\pm$ .106	& .532 	& $\pm$ .043           				\\
\addlinespace
			& RoBERTa-premises+conclusion		& \bf .896	& $\pm$ .025 		& \bf .888 	& $\pm$ .061    	& .887 & $\pm$ .048   & \bf .885 	& $\pm$ .029$^{\dagger\ddagger}$		\\
			& RoBERTa-premises+generated		& .889 	& $\pm$ .028      	& .879 	& $\pm$ .052       	& .883 	& $\pm$ .064       & .878 	& $\pm$ .030           				\\
			& RoBERTa-conclusion+generated		& .659 	& $\pm$ .026      	& .762 	& $\pm$ .030      	& .396 	& $\pm$ .092       & .571 	& $\pm$ .036           				\\
\addlinespace
			&\bf  RoBERTa-all					& \bf .896 	& $\pm$ .024 		& .886	& $\pm$ .045       	& \bf .889 	& $\pm$ .054   & \bf .885 	& $\pm$ .025$^{\dagger\ddagger}$	\\
\bottomrule
\end{tabular*}
 \caption{Results of argument sufficiency assessment: Accuracy as well as macro precision, recall, and F$_1$-score of all evaluated approaches, averaged over twenty 5-fold cross-validations. Significant gains over \citet{stab-gurevych-2017aae2} and the RoBERTa approach without structural enrichment are marked with $^{\dagger}$ and $^{\ddagger}$, respectively (computed using Wilcoxon signed-rank test, $p$-value .05). The human upper bound is obtained on a subset of 432 arguments.}
\label{fig:resulttable1}
\end{table*}

\section{Evaluation of Sufficiency Assessment}
\label{sec:assessment}

Using our conclusion generation model, BART-supervised, we finally study our hypothesis on sufficiency assessment by experimenting with different input variations and testing on the corpus of \newcite{stab2017recognizingsupportedarguments}. The first setting is by just using the RoBERTa model ({\em direct sufficiency assessment}) and the second by introducing structural knowledge and our generated conclusions ({\em indirect} sufficiency assessment).

\subsection{Direct Sufficiency Assessment}

First, we compare the ``base version'' of our approach, RoBERTa without additional structure annotations, to the {\em human upper bound} and the state of the art {\em CNN} of \citet{stab2017recognizingsupportedarguments}.%
\footnote{Note that the human upper bound of \newcite{stab2017recognizingsupportedarguments} was computed on a subset of 433 arguments annotated by three annotators only. Thus, it is only an approximation of the actual human performance. The human scores are based on pairwise comparisons of the three annotators.}

\paragraph{Results}

The upper part of Table \ref{fig:resulttable1} shows the direct assessment results. Our RoBERTa model significantly outperforms the CNN both on accuracy (.889 vs.\ .846) and on macro F$_1$ score (.876 vs.\ .831), the latter being an improvement of whole 4.5 points. Our model also performs almost on par with the human upper bound, meaning it is approximately at the level of human performance. This underlines the potential of pre-trained transformer models in argument quality assessment.

\subsection{Indirect Sufficiency Assessment}

As our conclusion generation model starts from structural annotations of a given argument, we systematically study the benefit of knowing the premises and the original conclusion, as well of having the generated conclusion. We consider the following eight input variants for the assessment:

\begin{itemize}
\setlength{\itemsep}{0pt}
\item 
\textit{RoBERTa-premises-only.} Use the full argument as input, but replace the ground-truth conclusion with an <\textbackslash unk> token.
\item 
\textit{RoBERTa-conclusion-only.} Use only the ground-truth conclusion as input.
\item 
\textit{RoBERTa-generated-only.} Use only the generated conclusion as input.
\item 
\textit{RoBERTa-premises+conclusion.} Use the full argument as input and highlight the ground-truth conclusion using <\textbackslash s> tokens.
\item 
\textit{RoBERTa-premises+generated.} Use the full argument as input, but replace the ground-truth conclusion with its generated counterpart. Highlight the latter using <\textbackslash s> tokens.
\item 
\textit{RoBERTa-conclusion+generated.} Use only the ground-truth and the generated conclusion as input, separated with a <\textbackslash s> token
\item 
\textit{RoBERTa-all.} Use the full argument as input, insert the highlight the generated conclusion after the ground-truth conclusion and highlight both together using <\textbackslash s> tokens.
\end{itemize}

\paragraph{Results}

The lower part of Table \ref{fig:resulttable1} shows that both {\em RoBERTa-premises+conclusion} and {\em RoBERTa-all} yield the best results overall, significantly outperforming our vanilla {\em RoBERTa} model by almost 1 point in terms of both accuracy (.896 vs.\ .889) and macro F$_1$-score (.885 vs.\ .876). 
The results suggest that using a generated conclusion for the argument does not really help, but also not hurts the model performance. This is also supported by {\em RoBERTa-premises-only}, which also matches the performance of {\em RoBERTa-premises+generated}. In general, however, bringing structural knowledge to the model gives a slight but significant improvement in sufficiency assessment. 

Looking at the weak performance of {\em RoBERTa-conclusion+generated} (macro F$_1$-score .571), we see that an opposition of the two conclusions alone is not enough four sufficiency assessment. Even though adding the generated one improves over having the ground-truth conclusion only, all variants that include the premises perform much better.

To better understand the role of premises and conclusions in sufficiency assessment, we trained the three {\em RoBERTa-<xy>-only} variants. The high performance of {\em RoBERTa-premises-only}  (macro F$_1$-score .875) clearly reveals that the conclusion is of almost no importance on the data of \newcite{stab2017recognizingsupportedarguments}, being not significantly worse than vanilla RoBERTa. The low scores of {\em RoBERTa-conclusion-only} further support this hypothesis, suggesting that the knowledge obtained from the conclusion can be inferred from the argument without its conclusion alone. This result is very insightful in that it displays that the currently available data barely enables a study of sufficiency assessment in terms of its actual definition. Instead, we suppose that models mainly learn a correlation between the nature and the quality of a given set of premises and the possible sufficiency evolving from this. In particular, students who provide ``good'' premises for an argument in their essays can also deliver an inferrable conclusion.

\section{Discussion}
\label{sec:discussion}

Our results suggest that large-scale pre-trained transformer models can help assess the quality of arguments, here their sufficiency. They even nearly matched human performance. However, an accurate understanding of the argument sufficiency task in terms of the actual definition of sufficiency seems barely possible on the available data.

Employing knowledge about argumentative structure can benefit sufficiency assessment, in line with findings on predicting essay-level argument quality \citep{wachsmuth2016using}. Our results suggest that there is at least some additional knowledge in an argument's conclusion that our model could not learn itself. However, we did not actually {\em mine} argumentative structure here, but we resorted to the human-annotated ground truth, which is usually not available in a real-world setting. Thus, the improvements obtained by the structure could vanish as soon as we resort to computational methods. We note, though, that we obtained state-of-the-art results also using RoBERTa on the plain text only.

Regarding the central hypothesis of this work, we study an example of a more task-aligned approach. However, our results show that answering the question of conclusion inferability by generating conclusions for an argument is difficult, as generated conclusions may be perceived as sufficient, likely, and novel, but may still not be unique. That is, for many premises, it may be possible to generated multiple sufficient conclusions, which naturally limits the impact of {\em quality assessment through generation}.

Consequently, although the task of sufficiency assessment appears to be solved on the given data, we argue for the need for more sophisticated corpora that better reflect the actual definition of the task, to ultimately allow studying whether approaches such as ours are needed in real-world applications.

\section{Conclusion}
\label{sec:conclusion}

In this work we have studied the task of argument sufficiency assessment based on auto-generated argument conclusions. According to our findings, traditional approaches can be improved by using large-scale pre-trained transformer models and by incorporating knowledge about argumentative structure. The effect of our proposed idea to leverage generation for the assessment turned out low though. However, this may likely be caused by the available data, wehere sufficiency seems to barely depend on the arguments' conclusions, thus preventing our and previous approaches from actually tackling the task as intended by its definition. 

In general, the insights of this paper lay the foundation for more task-oriented approaches towards the assessment of argument quality dimensions, that are tailored towards the properties in scope (here the relation between premises and conclusion). To adequately evaluate such approaches, also refined corpora may be needed.

\section*{Acknowledgments}

We thank Katharina Brennig, Simon Seidl, Abdullah Burak, Frederike Gurcke and Dr. Maurice Gurcke for their feedback. We gratefully acknowledge the computing time provided the described experiments by the Paderborn Center for Parallel Computing (PC$^2$). This project has been partially funded by the German Research Foundation~(DFG) within the project OASiS, project number 455913891, as part of the Priority Program ``Robust Argumentation Machines (RATIO)'' (SPP-1999).

\bibliography{argmining21-sufficiency-generation-lit}
\bibliographystyle{acl_natbib}


\end{document}